# CRC-HGD: A Histopathological Image Dataset for Grading Colorectal Cancer

Elham Amjadi[1], Amin Bahreini[1,2], Sayed Mohammad Hasan Emami[1], Sayyed Mohammadreza Hakimian[3], Alireza Fahim[1], Hojjatollah Rahimi[1], Hamidreza Bolhasani[4,*]

[1] *Poursina Hakim Digestive Diseases Research Center, Isfahan University of Medical Sciences, Isfahan, Iran*
[2] *Division of Transplant Services, Department of Surgery, SUNY Upstate Medical University, Syracuse, New York, USA*
[3] *Cancer Prevention Research Center, Isfahan University of Medical Sciences, Isfahan, Iran*
[4] *DataBioX Research, Tehran, Iran*

* **Corresponding Author:** hamidreza@databiox.com

**ABSTRACT**

Colorectal cancer (CRC) is the third most common cancer worldwide and the second leading cause of cancer-related deaths globally, with approximately 1,926,425 new cases and 904,019 deaths reported in 2022. Accurate histologic grading plays a critical role in prognosis and treatment planning for colorectal adenocarcinoma. In recent years, artificial intelligence and its subcategories, including machine learning and deep learning, have been increasingly employed for automated cancer detection and classification. An appropriate and well-organized dataset is the essential first step to achieve this goal. This paper introduces CRC-HGD, a histopathological microscopy image dataset of 1,914 images obtained from 214 colorectal adenocarcinoma patients (Grade I: 106, Grade II: 75, Grade III: 33). The specimens are H&E-stained colorectal tissue sections acquired at the Poursina Hakim Research Center of Isfahan University of Medical Sciences, Iran, diagnosed between 2014 and 2019, and graded according to the World Health Organization (WHO) criteria into three grades: well-differentiated (Grade I), moderately differentiated (Grade II), and poorly differentiated (Grade III). For each specimen, four magnification levels are provided: 4x, 10x, 20x, and 40x. The dataset is accessible via Mendeley Data (https://doi.org/10.17632/yfp5sfj47m.4) and at http://databiox.com, where the latest version is also available. The distinctive feature of this dataset is the provision of labeled specimens across all three differentiation grades at multiple magnification levels, enabling comprehensive computational analysis of colorectal cancer grading.

## 1. Introduction

Cancer is a serious public health issue worldwide. According to the Global Cancer Observatory (GLOBOCAN) 2022, colorectal cancer (CRC) is the third most common cancer in terms of incidence and the second leading cause of cancer-related mortality globally, with approximately 1,926,425 new cases and 904,019 deaths reported in 2022 [1, 2]. The global burden of CRC has been rising steadily, with the number of new cases increasing from 916,583 in 1990 to over 2,194,143 in 2021, underscoring the urgent need for improved early detection and treatment strategies [3].

More than 90% of colorectal carcinomas are adenocarcinomas originating from epithelial cells of the colorectal mucosa [4]. Conventional adenocarcinoma is characterized by glandular formation, which constitutes the basis for histologic tumor grading. According to the World Health Organization (WHO) criteria, colorectal adenocarcinomas are classified into three grades based on the proportion of gland formation: well-differentiated (Grade I, >95% gland formation), moderately differentiated (Grade II, 50–95% gland formation), and poorly differentiated (Grade III, <50% gland formation) [4, 5]. In practice, approximately 70% of colorectal adenocarcinomas are diagnosed as moderately differentiated, while well-differentiated and poorly differentiated tumors account for about 10% and 20% of cases, respectively [6].

Histologic grade is an independent prognostic factor for colorectal carcinoma. Poorly differentiated adenocarcinoma is associated with adverse clinical outcomes, with a 5-year survival rate of only 20% to 45.5%, compared to more favorable outcomes for well- and moderately differentiated subtypes [7, 8]. Therefore, accurate and early histologic grading is essential in selecting the appropriate treatment plan and improving patient survival.

Pathologists conventionally assess CRC grade through visual examination of hematoxylin and eosin (H&E) stained tissue sections under a microscope. This process is time-consuming and subject to significant inter-observer variability [9]. In recent years, remarkable efforts have been made to automate cancer detection and grading using artificial intelligence (AI), particularly machine learning (ML) and deep learning (DL) algorithms, which can assist pathologists in making faster and more accurate diagnoses [10–15]. Early and accurate AI-assisted grading has the potential to directly influence clinical decision-making, guide adjuvant chemotherapy selection, and ultimately improve patient survival outcomes. A well-organized and publicly available histopathological image dataset is the essential first step for training and evaluating such algorithms.

Reviewing existing datasets for colorectal cancer histopathology (Section 2), we found that although several datasets classify CRC tissue types or provide binary malignant/benign labels, no publicly available dataset specifically targets histologic grading (Grade I, II, III) of colorectal adenocarcinoma with multiple magnification levels per specimen. This gap motivated the creation of the CRC-HGD dataset presented in this work, collected from the Poursina Hakim Research Center of Isfahan University of Medical Sciences, Isfahan, Iran.

This paper is organized as follows: Related works are presented in Section 2. Section 3 describes the CRC-HGD dataset, including methods of image collection and dataset statistics. The value of the data and discussion are provided in Sections 4 and 5, respectively.

## 2. Related Works

Kather et al. [10] introduced the NCT-CRC-HE-100K dataset, one of the most widely used datasets in CRC computational pathology. It comprises 100,000 H&E-stained histological image patches extracted from 86 colorectal cancer slides, spanning nine tissue classes including adipose tissue, background, debris, lymphocytes, mucus, smooth muscle, normal colon mucosa, cancer-associated stroma, and colorectal adenocarcinoma epithelium. Images have a resolution of 224 × 224 pixels at 0.5 μm per pixel. While this dataset supports multi-class tissue classification, it does not provide histologic grade labels.

Kather et al. [11] also released an earlier texture dataset (Kather_texture_2016) consisting of 5,000 histological images of 150 × 150 pixels, categorizing eight tissue types of human colorectal cancer. This dataset has been extensively used for texture-based classification benchmarks [12, 13], but does not address tumor differentiation grades.

Shi et al. [14] introduced EBHI-Seg, the first dataset designed for multi-class segmentation of colorectal biopsy H&E images. It covers six histological differentiation categories: Normal, Hyperplasia polyp, Low-grade dysplasia, High-grade dysplasia, Serrated adenoma, and Adenocarcinoma, with pixel-level ground truth masks. While this dataset provides a broad histological categorization, it does not follow the WHO three-tier grading system for colorectal adenocarcinoma.

Liang et al. [15] proposed the HCCANet model for histopathological image grading of colorectal cancer using a CNN with a multichannel fusion attention mechanism, evaluated on three differentiation levels and achieving an overall accuracy of 87.3%. However, the underlying dataset was institution-specific and not publicly released, limiting reproducibility.

Yoshida et al. [16] trained deep learning models for classifying poorly differentiated colorectal adenocarcinoma in whole slide images (WSIs) using transfer learning. Their dataset was also institution-specific and not publicly released.

To the best of our knowledge, no publicly available histopathological microscopy image dataset specifically targeting the WHO three-tier grading of colorectal adenocarcinoma with multiple magnification levels per specimen currently exists. This gap was the primary motivation for creating the CRC-HGD dataset.

## 3. The CRC-HGD Dataset

CRC-HGD (Colorectal Cancer Histopathological Grading Dataset) is a histopathological microscopy image dataset of colorectal adenocarcinoma patients prepared for grade classification. The dataset includes a total of 1,914 images in RGB format, JPEG type, at a resolution of 800 × 800 pixels and 96 DPI. Table 1 presents the technical specifications of the dataset.

The naming convention of the files in the dataset is based on the format provided in Table 2. An example filename is: 01_CRC_G1_16130_4x_1. The last segment following the magnification level (e.g., "_1") indicates the image index at that magnification level for the given specimen, since multiple images may be acquired per specimen at the same magnification.

The specimens are colorectal tissue sections stained with H&E, taken from 214 patients (Grade I: 106, Grade II: 75, Grade III: 33) who were diagnosed between 2014 and 2019 in the Poursina Hakim Research Center of Isfahan University of Medical Sciences, Isfahan, Iran. This dataset is published and accessible via Mendeley Data (http://doi.org/10.17632/yfp5sfj47m.4) and is also available at http://databiox.com. For the latest updates and additional information related to this dataset, readers are encouraged to visit http://databiox.com.

For each specimen, four levels of magnification are provided: 4x, 10x, 20x, and 40x. Based on the pathologist's assessment, in some cases more than one image from a specific magnification level is captured per specimen; this is reflected in the last index segment of the filename. Labeling was performed according to WHO criteria for colorectal adenocarcinoma grading based on the proportion of gland-forming structures: Grade I (well-differentiated, >95% gland formation), Grade II (moderately differentiated, 50–95% gland formation), and Grade III (poorly differentiated, <50% gland formation) [4, 5]. The dataset also includes 8 images of normal colorectal tissue (2 per magnification level) and 7 mixed normal-tumoral images. Representative H&E-stained images at all four magnification levels for each grade are shown in Figure 2.

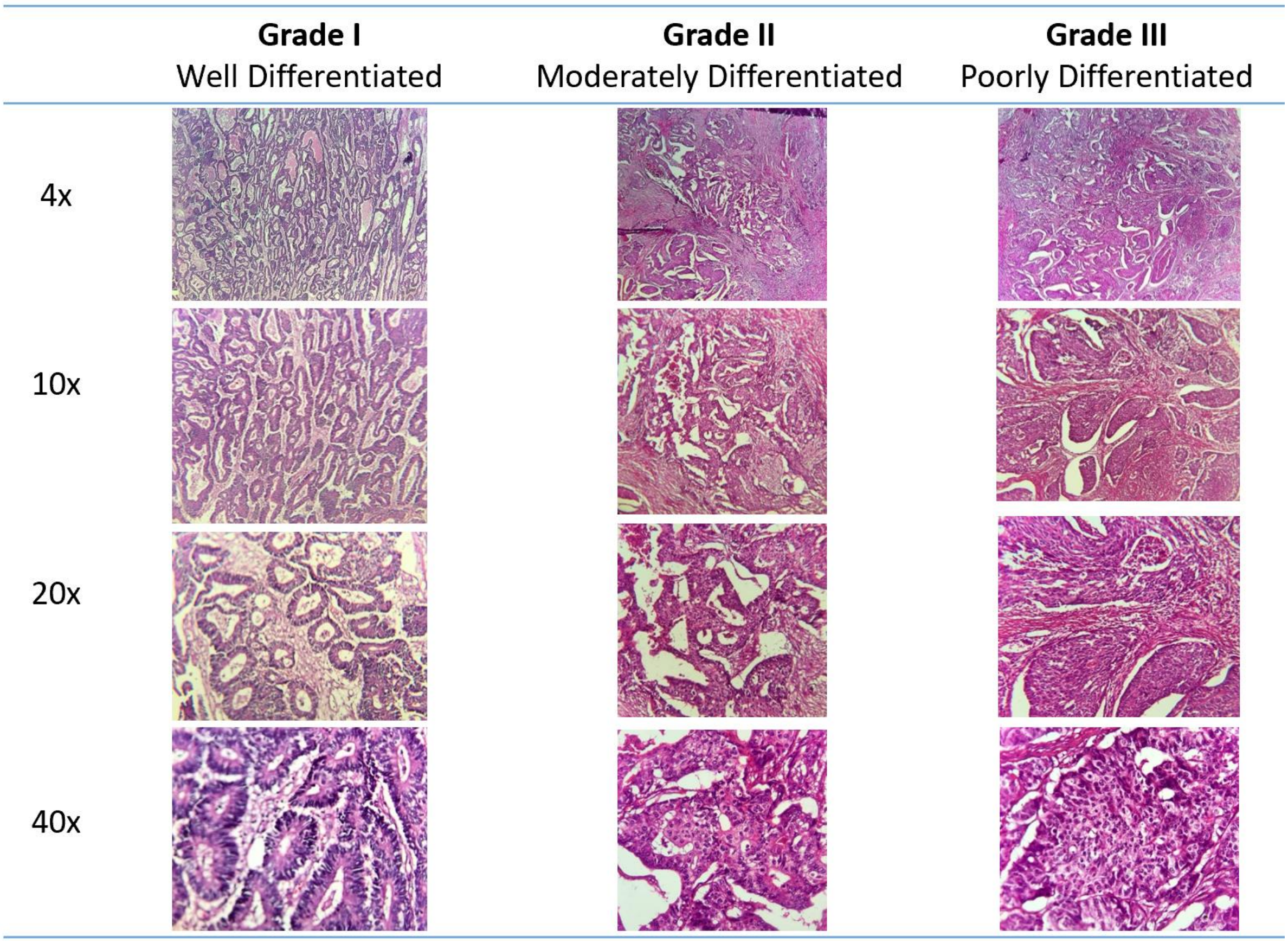

*Figure 2. Representative H&E-stained colorectal tissue images from the CRC-HGD dataset at four magnification levels (4x, 10x, 20x, 40x) for Grade I (well-differentiated), Grade II (moderately differentiated), and Grade III (poorly differentiated) adenocarcinoma.*

What distinguishes this dataset from existing ones is the provision of histologically graded specimens across all three WHO differentiation levels, each imaged at four magnification levels. Overall, five essential steps were taken to prepare this dataset:

1. Selecting patients diagnosed with colorectal adenocarcinoma
2. Collecting H&E stained tissue slides from pathology archives
3. Labeling specimens based on WHO histologic grading criteria (Grade I, II, III)
4. Microscopy imaging at four magnification levels (4x, 10x, 20x, 40x)
5. Organizing and publishing the dataset

**Table 1**

Technical specifications of the CRC-HGD dataset.

| # | Dataset Specification | Value |
|---|---|---|
| 1 | Number of images | 1,914 |
| 2 | Number of patients | 214 (Grade I: 106 \| Grade II: 75 \| Grade III: 33) |
| 3 | Image format | RGB |
| 4 | File type | JPEG |
| 5 | Resolution | 800 × 800 pixels |
| 6 | DPI | 96 |
| 7 | Bit depth | 24 |
| 8 | Magnification levels | 4x, 10x, 20x, 40x |
| 9 | Staining | Hematoxylin and Eosin (H&E) |
| 10 | Grading system | WHO three-tier (Grade I, II, III) |
| 11 | Dataset URL | http://databiox.com<br>https://doi.org/10.17632/yfp5sfj47m.4 |

**Table 2**

Naming structure of the files in the CRC-HGD dataset.

| Part | Description | Sample 1 | Sample 2 |
|---|---|---|---|
| 1 | Sample number | 01 | 103 |
| 2 | Cancer type | CRC | CRC |
| 3 | Grade label | G1 | G1 |
| 4 | Pathology archive number | 16130 | 230472 |
| 5 | Magnification level | 4x | 4x |
| 6 | Image index at magnification | 1 | 1 |

**Table 3**

Detailed dataset information based on specimens per grade and magnification level.

| Category | Patients | 4x | 10x | 20x | 40x | Total |
|---|---|---|---|---|---|---|
| Grade I (Well-diff.) | 106 | 106 | 244 | 251 | 259 | 860 |
| Grade II (Mod.-diff.) | 75 | 75 | 210 | 210 | 217 | 712 |
| Grade III (Poorly-diff.) | 33 | 33 | 97 | 99 | 98 | 327 |
| Normal | — | 2 | 2 | 2 | 2 | 8 |
| Mixed Normal/Tumoral | — | — | — | — | — | 7 |
| Total | 214 | 216 | 553 | 562 | 576 | **1,914** |

Detailed information about the quantity of specimens based on their grades and magnification levels is presented in Table 3 and Figures 1 and 3–7. Figure 1 shows the number of patients and total images per grade. Figures 3, 4, and 5 display the distribution of images per magnification level for Grade I, Grade II, and Grade III, respectively. Figure 6 shows the proportional distribution of patients and images by category. Figure 7 presents the stacked image count per magnification level across all grades.

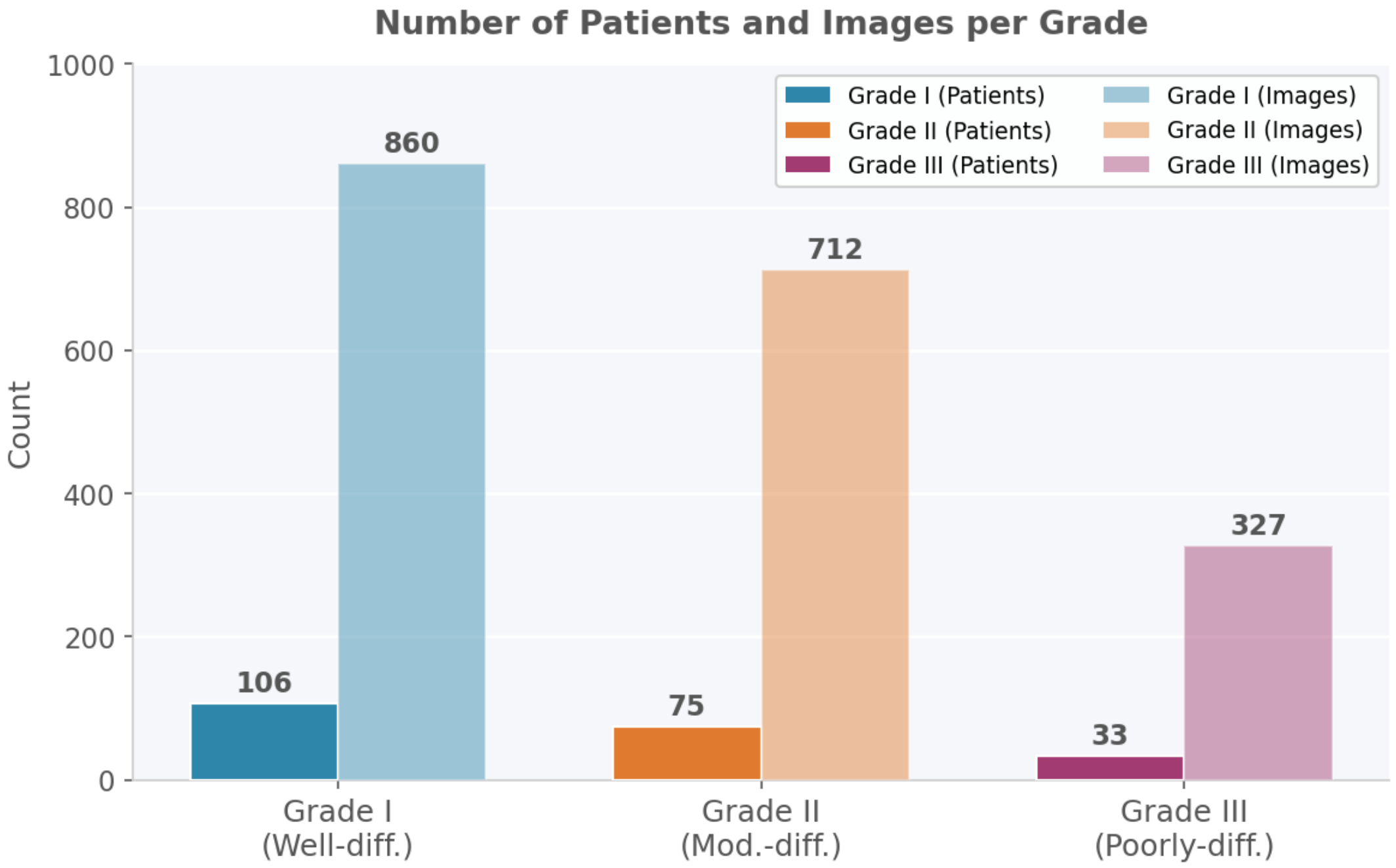

*Figure 1. Number of patients and total images per grade in the CRC-HGD dataset (Grade I: 106 patients / 860 images; Grade II: 75 patients / 712 images; Grade III: 33 patients / 327 images).*

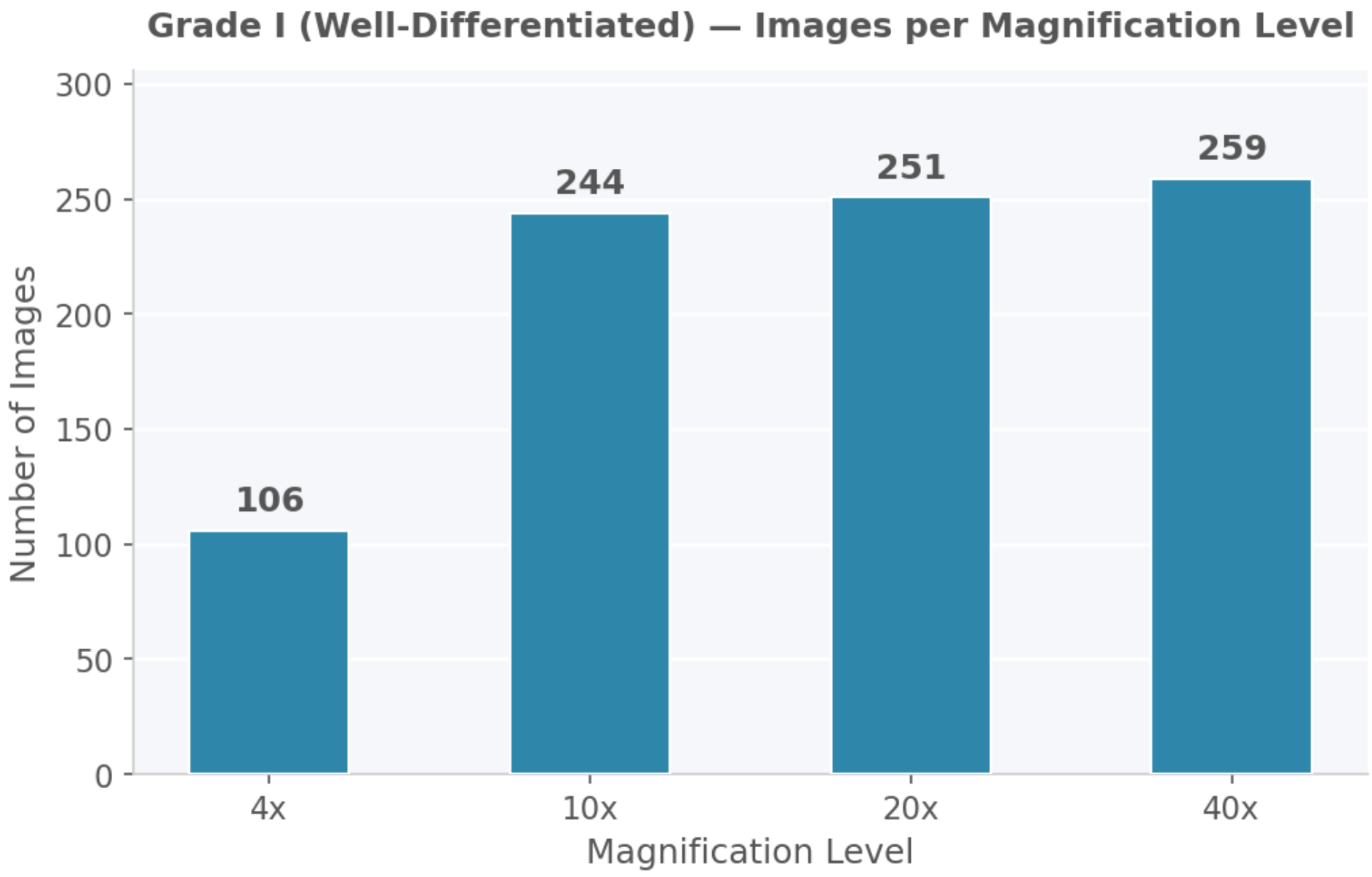

*Figure 3. Distribution of Grade I (well-differentiated) images per magnification level: 4x=106, 10x=244, 20x=251, 40x=259.*

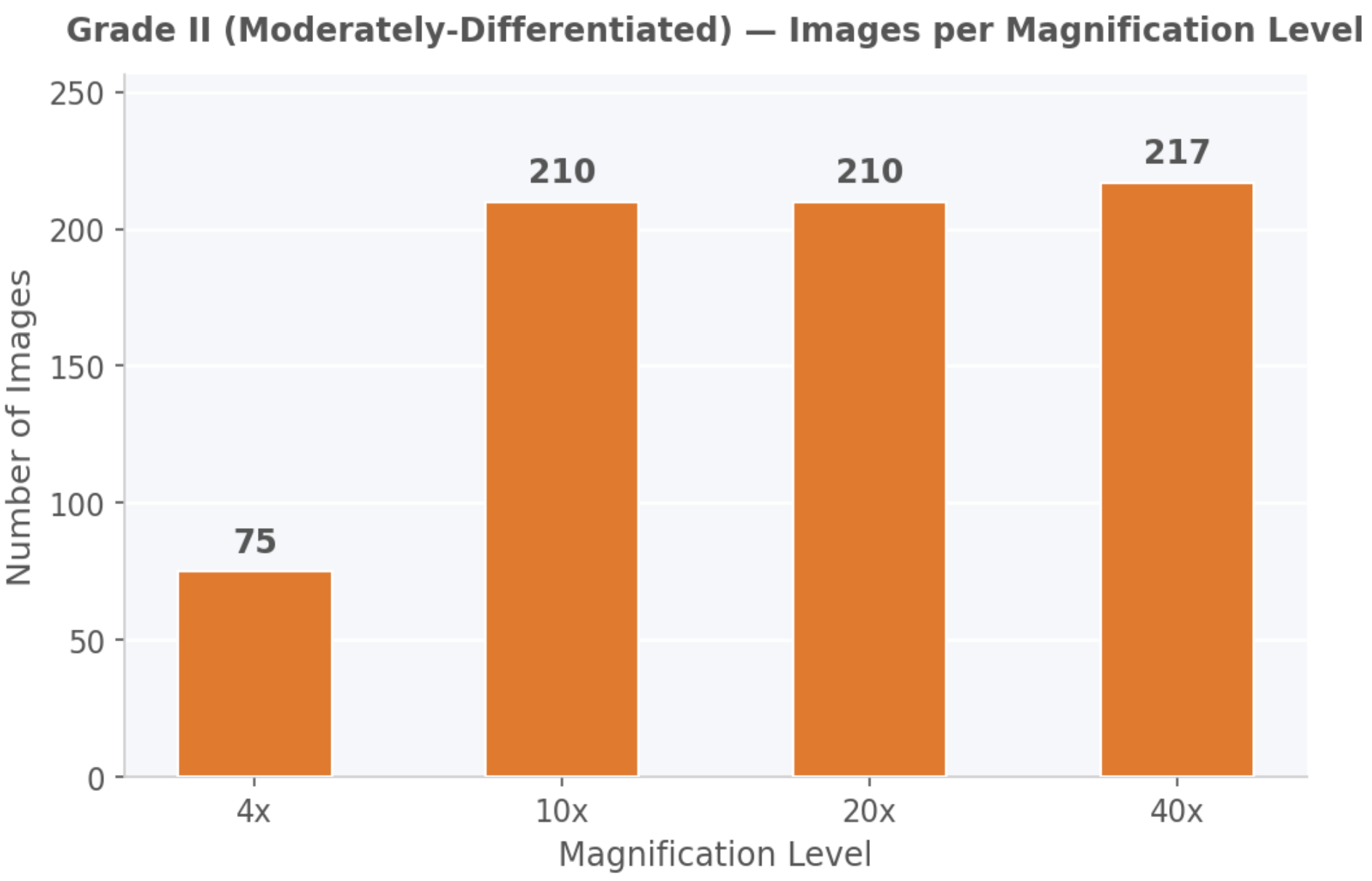

*Figure 4. Distribution of Grade II (moderately differentiated) images per magnification level: 4x=75, 10x=210, 20x=210, 40x=217.*

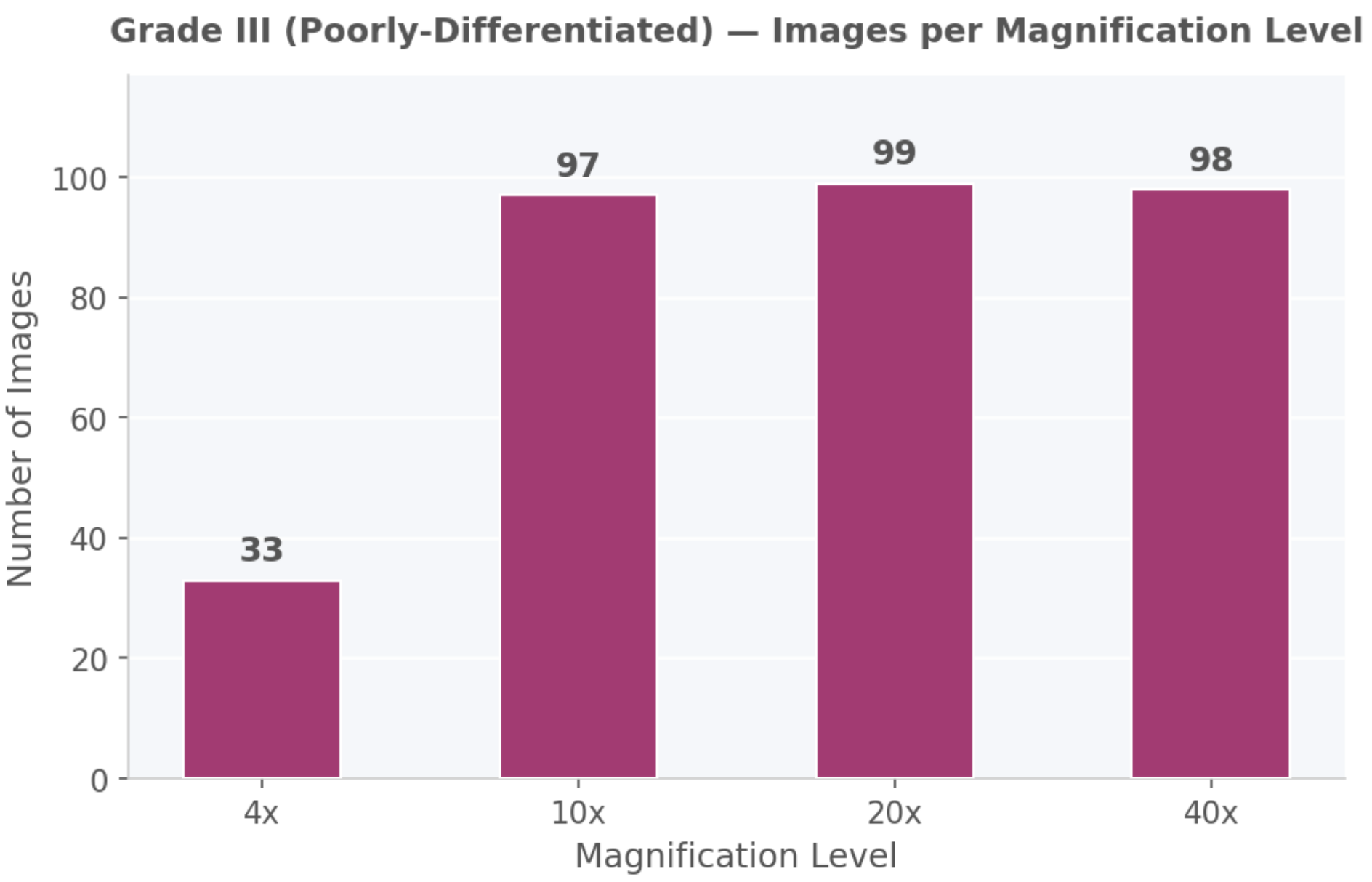

*Figure 5. Distribution of Grade III (poorly differentiated) images per magnification level: 4x=33, 10x=97, 20x=99, 40x=98.*

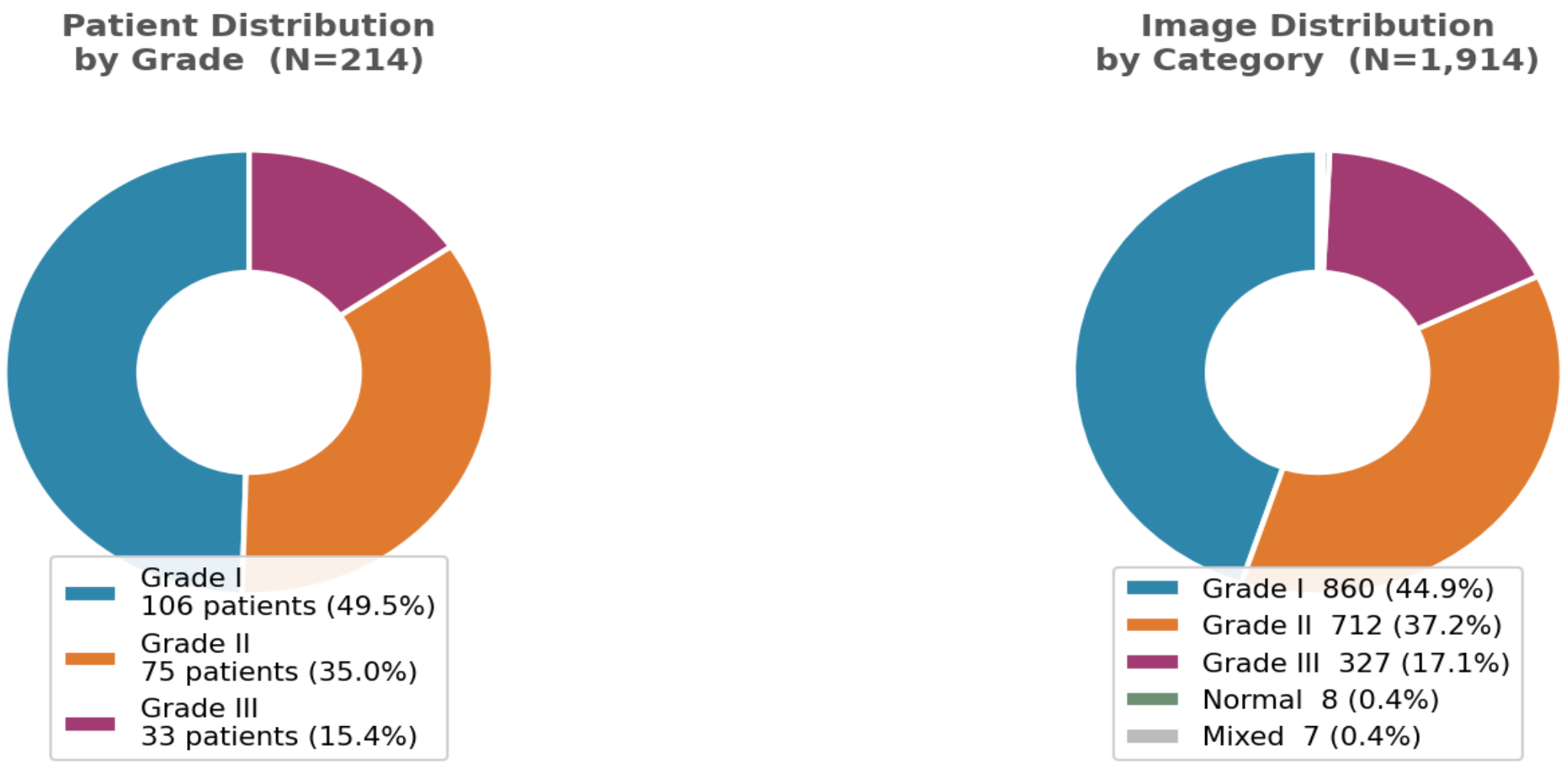

*Figure 6. Proportional distribution of (left) patients and (right) images across all categories in the CRC-HGD dataset.*

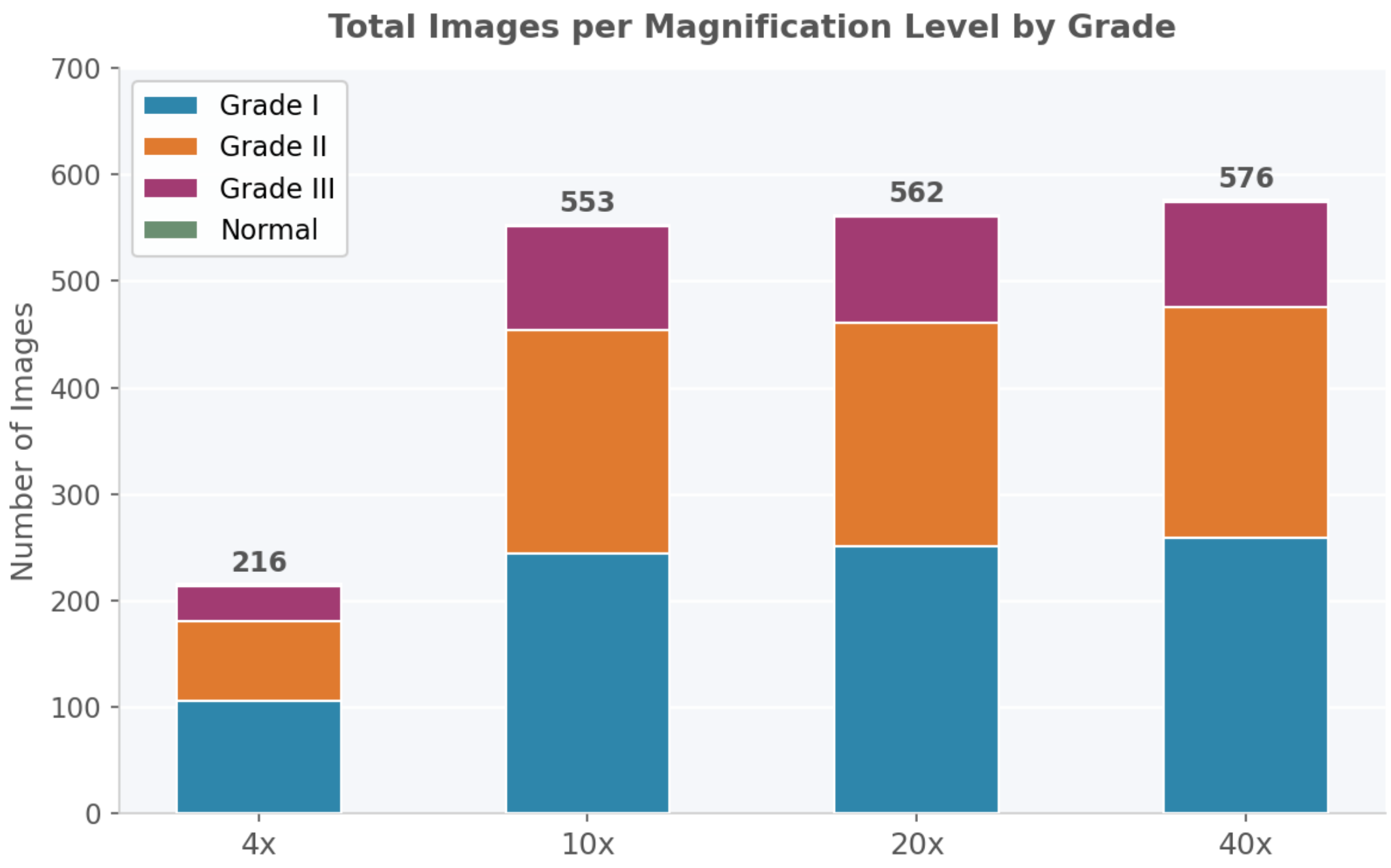

*Figure 7. Stacked bar chart showing total image counts per magnification level stratified by grade (4x=216, 10x=553, 20x=562, 40x=576).*

## 4. Value of the Data

- The CRC-HGD is a comprehensive histopathological image dataset specifically designed for colorectal adenocarcinoma histologic grading, filling a critical gap in the existing publicly available CRC datasets.
- All 1,914 images are labeled based on the WHO three-tier grading system (Grade I: well-differentiated; Grade II: moderately differentiated; Grade III: poorly differentiated) and magnification level, enabling direct use in supervised learning pipelines.
- The dataset provides images from 214 patients across all three differentiation grades at four magnification levels (4x, 10x, 20x, 40x), enabling multi-scale computational analysis of tumor morphology.

- This dataset can serve as a training and evaluation benchmark for machine learning and deep learning models targeting automated CRC grade detection, thereby supporting AI-assisted pathology and improving diagnostic reproducibility.
- AI-assisted grading based on this dataset can contribute to more accurate and reproducible cancer diagnoses, guide adjuvant chemotherapy selection, assist in treatment planning, and ultimately improve patient outcomes through early identification of high-grade, aggressive tumors.
- The inclusion of normal tissue images (8 images) and mixed normal-tumoral specimens (7 images) provides additional contrast for model training and supports tumor boundary analysis.
- The dataset is publicly accessible at http://databiox.com and via Mendeley Data (http://doi.org/10.17632/yfp5sfj47m.4), facilitating reproducibility and open science in computational pathology. For the latest updates and additional information, please visit http://databiox.com.

## 5. Discussion

Recently, significant efforts have been made to predict and detect colorectal cancer using artificial intelligence and its subcategories including machine learning and deep learning. These methods have demonstrated promising results in tissue classification and tumor detection; however, the specific task of histologic grade classification of colorectal adenocarcinoma has received comparatively less attention, in part due to the lack of publicly available, grade-labeled datasets. A well-organized dataset is an essential prerequisite for training, validating, and benchmarking such algorithms.

Histologic grading is a substantial prognostic factor in colorectal adenocarcinoma. Reviewing major related works in the field [10–16] revealed that while several datasets exist for CRC tissue classification, no publicly available dataset specifically targets WHO-based histologic grading (Grade I, II, III) with multiple magnification levels per specimen. This gap was the primary motivation for preparing CRC-HGD.

CRC-HGD includes 1,914 histopathological images in RGB JPEG format (800 × 800 pixels, 96 DPI), taken from 214 patients (Grade I: 106, Grade II: 75, Grade III: 33) diagnosed with colorectal adenocarcinoma between 2014 and 2019 at the Poursina Hakim Research Center of Isfahan University of Medical Sciences, Isfahan, Iran. For each specimen, four levels of magnification (4x, 10x, 20x, and 40x) are provided. All tumor images are labeled based on the WHO histologic grade, and the dataset also includes 8 normal tissue images and 7 mixed normal-tumoral images to support more comprehensive model training. The dataset is accessible at http://databiox.com and via Mendeley Data (http://doi.org/10.17632/yfp5sfj47m.4). For the latest updates and additional information related to this dataset, readers are encouraged to visit http://databiox.com.

Accurate AI-assisted grading of colorectal adenocarcinoma has direct clinical implications. Early and reliable identification of high-grade (Grade III, poorly differentiated) tumors can guide clinicians toward more aggressive treatment strategies and closer postoperative monitoring, ultimately contributing to improved patient survival. The availability of a standardized, publicly accessible dataset such as CRC-HGD is expected to catalyze further research in computational pathology and support the development of robust, reproducible AI-based grading tools for colorectal cancer.

## Author Biographies

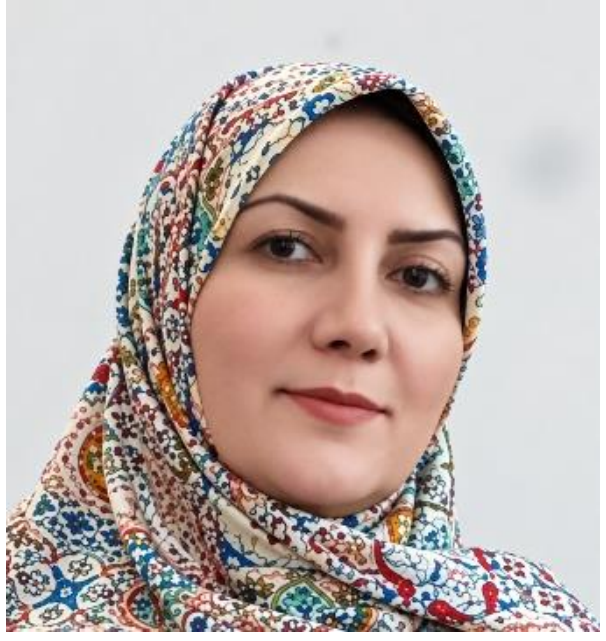

**Elham Amjadi, MD** is a Pathologist with expertise in gastrointestinal and breast pathology. She received her medical degree and completed her residency in pathology at Isfahan University of Medical Sciences. Her research interests include histopathological grading of gastrointestinal malignancies and AI-assisted pathology. She has been actively involved in the development of computational pathology datasets and AI-based diagnostic tools.

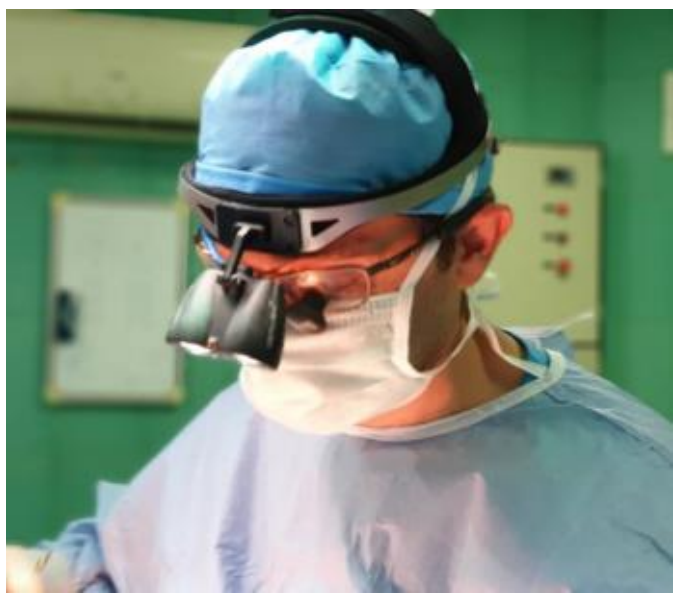

**Amin Bahreini, MD** is a Transplant and Hepatobiliary Surgeon at the Division of Transplant Services, Department of Surgery, SUNY Upstate Medical University, Syracuse, New York, USA. He completed fellowship training in multi-organ abdominal transplant surgery and hepatobiliary surgery, with advanced fellowships in liver and kidney transplantation. His research interests include transplant outcomes, organ preservation, and the application of data science in surgical decision-making.

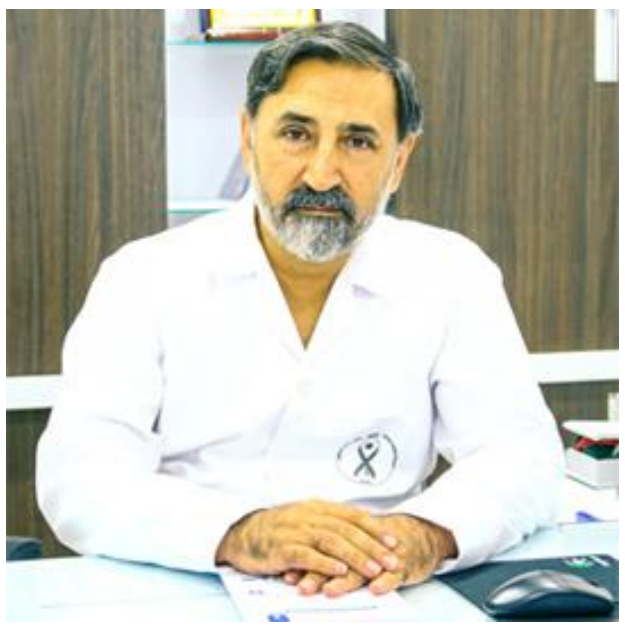

**Sayed Mohammad Hasan Emami, MD** is a Professor of Gastroenterology and Full Professor at Isfahan University of Medical Sciences, Isfahan, Iran. He is affiliated with the Poursina Hakim Digestive Diseases Research Center (PDDRC), where he has made significant contributions to gastrointestinal oncology and digestive disease research. His research interests include colorectal cancer, inflammatory bowel disease, and clinical outcomes in gastrointestinal malignancies.

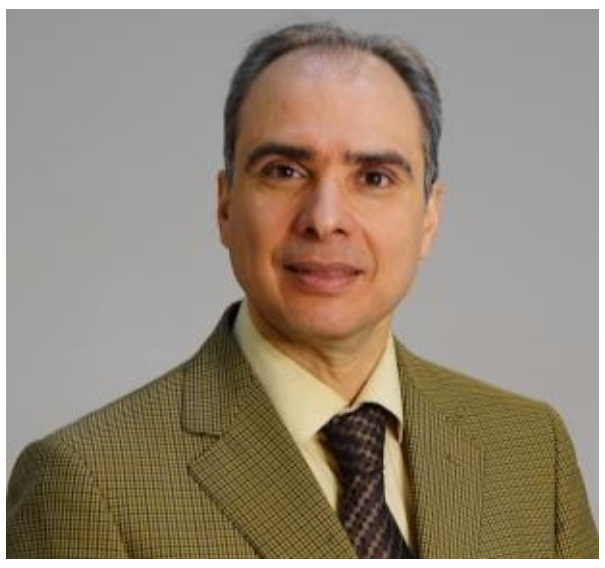

**Sayyed Mohammadreza Hakimian, MD** is a General Surgeon with fellowship training in surgical oncology, affiliated with the Cancer Prevention Research Center, Isfahan University of Medical Sciences, Isfahan, Iran. His clinical and research interests include gastrointestinal cancer surgery, breast cancer surgery, and oncological outcomes.

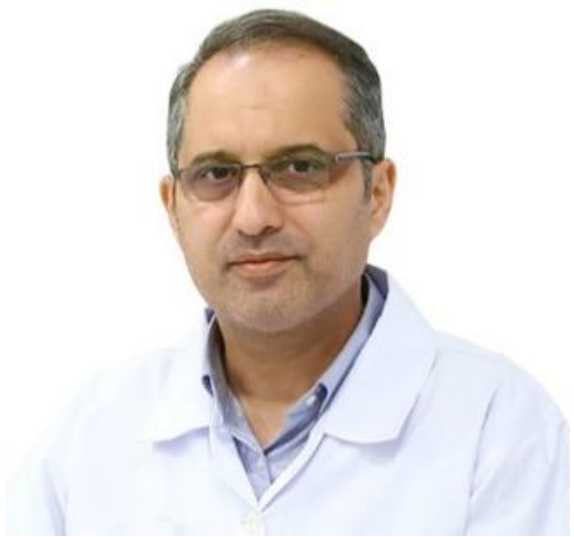

**Alireza Fahim, MD** is a Gastroenterology Subspecialist affiliated with the Poursina Hakim Digestive Diseases Research Center, Isfahan University of Medical Sciences, Isfahan, Iran. His clinical expertise and research focus on endoscopic diagnosis, gastrointestinal oncology, and histopathological evaluation of colorectal lesions.

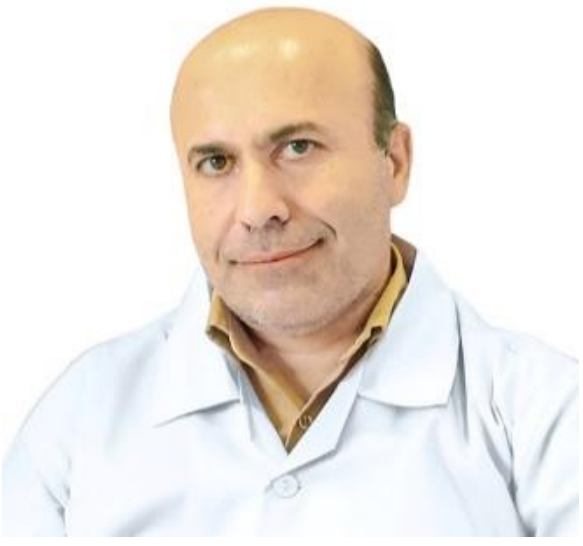

**Hojjatollah Rahimi, MD** is a Gastroenterology Subspecialist affiliated with the Poursina Hakim Digestive Diseases Research Center, Isfahan University of Medical Sciences, Isfahan, Iran. His clinical and research focus includes gastrointestinal oncology, colorectal cancer diagnosis, and endoscopic evaluation of digestive diseases.

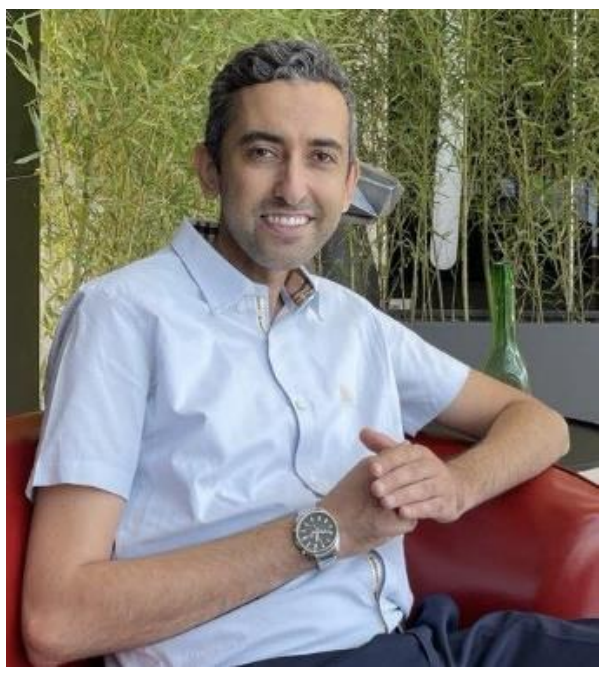

**Hamidreza Bolhasani, PhD** is an AI/ML Researcher, Founder and Chief Data Scientist at DataBioX (https://databiox.com), and Accepted Postdoctoral Researcher at Harvard Medical School (2022). He holds a PhD in Computer Engineering. His research focuses on artificial intelligence, machine learning, deep learning, computer vision, and digital pathology. He has developed multiple AI-based biomedical datasets and tools for colorectal lesion detection.